\pdfoutput=1
\def\year{2023}\relax
\documentclass[letterpaper]{article} 
\usepackage{aaai23}  
\usepackage{times}  
\usepackage{helvet}  
\usepackage{courier}  
\usepackage[hyphens]{url}  
\usepackage{graphicx} 
\usepackage{natbib}  
\usepackage{caption} 
\DeclareCaptionStyle{ruled}{labelfont=normalfont,labelsep=colon,strut=off} 
\usepackage{multirow}
\usepackage[T1]{fontenc}
\DeclareTextSymbolDefault{\dh}{T1}
\usepackage{algorithm}
\usepackage{algorithmic}
\usepackage[table,xcdraw]{xcolor}
\usepackage{newfloat}
\usepackage{listings}
\floatstyle{ruled}
\newfloat{listing}{tb}{lst}{}
\floatname{listing}{Listing}
\title{VaxxHesitancy: A Dataset for Studying Hesitancy towards COVID-19 Vaccination on Twitter}
\author{
    Yida Mu,
    Mali Jin,
    Charlie Grimshaw,
    Carolina Scarton,
    Kalina Bontcheva,
     Xingyi Song
}
\affiliations{
    \textsuperscript{\rm }Department of Computer Science, The University of Sheffield, UK\\

    {\{y.mu, m.jin, cgrimshaw1,  c.scarton, k.bontcheva, x.song\}@sheffield.ac.uk} \\
}

\begin{document}

\maketitle

\begin{abstract}
Vaccine hesitancy has been a common concern, probably since vaccines were created and, with the popularisation of social media, people started to express their concerns about vaccines online  alongside those posting pro- and anti-vaccine content. Predictably, since the first mentions of a COVID-19 vaccine, social media users posted about their fears and concerns or about their support and belief into the effectiveness of these rapidly developing vaccines. Identifying and understanding the reasons behind public hesitancy towards COVID-19 vaccines is important for policy markers that need to develop actions to better inform the population with the aim of increasing vaccine take-up. In the case of COVID-19, where the fast development of the vaccines was mirrored closely by growth in anti-vaxx disinformation, automatic means of detecting citizen attitudes towards vaccination became necessary. This is an important computational social sciences task that requires data analysis in order to gain in-depth understanding of the phenomena at hand. Annotated data is also necessary for training data-driven models for more nuanced analysis of attitudes towards vaccination. To this end, we created a new collection of over 3,101 tweets annotated with users' attitudes towards COVID-19 vaccination (stance). Besides, we also develop a domain-specific language model (VaxxBERT) that achieves the best predictive performance (73.0 accuracy and 69.3 F1-score) as compared to a robust set of baselines. To the best of our knowledge, these are the first dataset and model that model vaccine hesitancy as a category distinct from pro- and anti-vaccine stance.
\end{abstract}

\section{Introduction}
When the first COVID-19 vaccine was publicly administered in the UK in the end of 2020,\footnote{\url{https://www.england.nhs.uk/2020/12/landmark-moment-as-first-nhs-patient-receives-covid-19-vaccination/}} most counties also begun to promote vaccination to the public. A high vaccination uptake is considered the most reliable and effective way to contain the COVID-19 pandemic and protect high-risk groups, since these vaccines have been shown to prevent serious illnesses caused by the SARS-CoV-2 virus.\footnote{\url{https://www.pfizer.com/news/press-release/press-release-detail/pfizer-and-biontech-confirm-high-efficacy-and-no-serious}}

Nevertheless, despite the significant impact of COVID-19 on peoples lives, including restrictions on free movement and the requirement to use masks, the vaccines raised also concerns in a significant number of citizens worldwide with corresponding negative impact on countries' vaccination rates.
One of the reasons is vaccine disinformation online  \citep{muric2021covid,gisondi2022deadly}. 
Examples can be found easily in social media, where an "anti-vaxxer" (i.e. a user promoting narratives against vaccination) posts disinformation (e.g., false statistics or conspiracies) about vaccines to discourage their followers from getting vaccinated \citep{loomba2021measuring,jennings2021lack}. Common topics of such disinformation are usually related to politics and religion in an attempt to deliberately deceive the public \citep{rzymski2021strategies,van2021inoculating}. Some real-world examples appear in Table \ref{tab:examples} (see Tweets 2 \& 3) and there are already research datasets  \citep{cotfas2021longest,chen2022multilingual} that have collected and categorized such anti-vaxx posts.

On the other hand, medical professionals and some governments\footnote{For example, the UK one: \url{https://www.ons.gov.uk/peoplepopulationandcommunity/healthandsocialcare/healthandwellbeing}} recognised that some citizens can be hesitant towards vaccination, but through suitably targeted public health information campaigns they might be persuaded to take it up. Crucially also vaccine hesitant citizens are usually not actively trying to stop others from being vaccinated by spreading anti-vaxx content. The most common reasons behind hesitancy tend to arise from limited understanding of the way COVID-19 vaccines work, e.g. new biotechnologies such as mRNA, side effects, natural immunity \citep{poddar2022winds,CAVES}.

\begin{table*}[htp]
\centering
\resizebox{\textwidth}{!}{%
\begin{tabular}{|l|l|l|}
\hline
\textbf{\#} &
  \multicolumn{1}{c|}{\textbf{Definitions}} &
  \multicolumn{1}{c|}{\textbf{Examples}} \\ \hline
\textbf{Pro} &
  \textit{\begin{tabular}[c]{@{}l@{}}The tweet expresses opinions and actions \\ supporting COVID-19 vaccination use.\end{tabular}} &
  \textit{\begin{tabular}[c]{@{}l@{}}T1: Coronavirus vaccine developed by Oxford University \\ appears safe and trains the immune system. HTTPURL\end{tabular}} \\ \hline
\textbf{Anti} &
  \textit{\begin{tabular}[c]{@{}l@{}}The tweet expresses opinions and actions \\ against COVID-19 vaccination, with the \\ aim of, persuading others to refuse \\ vaccination. Please be aware that tweets\\  expressing the person’s own intention \\ to refuse vaccination themselves belong \\ to the “Vaccine Hesitant” category.\end{tabular}} &
  \textit{\begin{tabular}[c]{@{}l@{}}T2: Will it alter DNA \& remote control ppl thru 5g ? Yes\\ \\ T3: @USER BREAKING NEWS: \#Pfizervaccine Impact \\ on Fertility...\\ \\ T4: the vaccine was a set up for the micro chips,\\ Trump going be president again. After Bidens 4yrs up\end{tabular}} \\ \hline
\textbf{Hesitancy} &
  \textit{\begin{tabular}[c]{@{}l@{}}The tweet is centered on the person’s \\ intention to delay and/or refuse the \\ vaccine and is often in first person.\end{tabular}} &
  \textit{\begin{tabular}[c]{@{}l@{}}T5: Why is there even a vaccine for an illness with a such \\ high survival rate?\\ T6: \textit{I think it came out about 8 months after the pandemic}\\ \textit{hit and I was like absolutely no,}  \textit{that's too quick. I would} \\ \textit{I would not be happy with that.} \end{tabular}} \\ \hline
\textbf{Irrelevant} &
  \textit{\begin{tabular}[c]{@{}l@{}}Any tweet that does not express stance\\  towards the COVID-19 vaccine, e.g., \\ tweets about other kinds of vaccines \\ or tweets primarily about COVID-19 \\ or other aspects of the pandemic.\end{tabular}} &
  \textit{T7: @USER What are the recent news of COVID vaccine?} \\ \hline
\end{tabular}%
}
\caption{Category definitions and tweet examples.}
\label{tab:examples}
\end{table*}

\begin{table*}[!t]
\small
\centering
\resizebox{\textwidth}{!}{%
\begin{tabular}{|c|c|c|c|c|}
\hline
\textbf{Dataset} &
  \multicolumn{1}{c|}{\textbf{Time}} & 
  \textbf{Tweets} &
  \multicolumn{1}{c|}{\textbf{Labels}} &
  \textbf{Language} \\ \hline
\citet{cotfas2021longest} & Nov 2020 $\sim$Dec 2020         & 2,792   & in favor, against, neutral      & en                 \\ \hline
\citet{poddar2022winds} & Jan  2020 $\sim$March 2021         & 1,700  & in favor, against, neutral       & en                 \\ \hline
\citet{chen2022multilingual}   & Jan  2020 $\sim$March 2021  & 17,934  & positive, negative, neutral, off-topic       & fr, de, etc. \\ \hline

\citet{di2022vaccineu}   & Nov  2020 $\sim$ June 2021  & 3,000 & pro, anti, neutral, out-of-context       & es, de, it. \\ \hline

\citet{delcea2022new} & July 2021 $\sim$ Aug 2021    & 4,636 & in favor, against, neutral       & en                 \\ \hline
\textbf{VaxxHesitancy}   & \textbf{Nov 2020 $\sim$April 2022}         & 3,101  & \textbf{pro, anti, hesitancy, irrelevant}   & en                 \\ \hline
\end{tabular}%
}
\caption{Differences in specifications between previous datasets v.s. ours. Note that the `off-topic' and `out-of-context' labels denote the tweet is irrelevant to COVID-19 vaccines or vaccination in \citet{chen2022multilingual} and \citet{di2022vaccineu}; while the `irrelevant' label in our dataset is the same as the `Neutral' label in existing datasets, i.e., no explicit attitude or intention is expressed.}
\label{tab:related_work}
\end{table*}

Although previous work has also classified such posts (e.g., Tweets 5 \& 6 in Table ~\ref{tab:examples}) as anti-vaxx \citep{cotfas2021longest}, we argue that the  hesitant stance in these posts is different from that of anti-vaxx posts, which are spreading false information. The ability to monitor and understand citizens' unanswered questions and concerns about vaccines is critical for policymakers and vaccine developers, as it enables them to undertake better targeted information campaigns and other actions aimed at improving trust in government policies. In addition, the separation of vaccine hesitant posts from anti-vaxx disinformation benefits the content moderation and debunking efforts of fact-checkers and social media platforms.

Specifically within the field of computational social science, prior data-driven research on user attitudes towards COVID-19 vaccination \citep{muller2019crowdbreaks,cotfas2021longest,poddar2022winds} has considered vaccine stance of social media posts as a three-way classification task, namely: pro-vaxx, anti-vaxx and neutral.\footnote{Some of these categories may be named differently by different researchers. For consistency, we use pro-vaxx to encompass all similar labels used previously, e.g., \textit{in favor}, \textit{pro}, \textit{positive}.}  

However, we argue that a more nuanced understanding of online debates on vaccination is needed, especially with respect to detecting and tracking the concerns voiced by vaccine hesitant citizens. To this end, we introduce a new, separate stance category of \textbf{vaccine hesitancy} which is centered on the person’s intention to delay or refuse the vaccine and is usually expressed in first person.

To the best of our knowledge, we are the first to consider COVID-19 \textbf{vaccine hesitancy} as a separate stance category. 
The main contributions of this paper include:
\begin{itemize}
    \item We re-frame the task of vaccine stance classification into a \textbf{four-way classification} setup, by considering vaccine hesitancy as a separate stance category.
    \item For this new, more nuanced classification task, we create a \textbf{publicly available dataset}\footnote{https://github.com/GateNLP/VaxxHesitancy} of over 3,100 COVID-19 vaccine-related tweets labelled as belong to one of four stance categories: \textit{pro-vaxx}, \textit{anti-vaxx}, \textit{vaxx-hesitant}, or \textit{irrelevant}, which also spans the longest time period. The characteristics of our dataset are compared to those existing three-way classification ones in Table ~\ref{tab:related_work}. 
    \item We make available a \textbf{domain-specific language models (VaxxBERT)}, pre-trained on 175 million unlabelled COVID-19 vaccine-related tweets.

    \item We also provide a \textbf{linguistic analysis}, which highlights the difference in language patterns between \textit{anti-vaxx} and \textit{vaxx-hesitant} posts.
\end{itemize}

\section{Related Work}
The development of social networks brought an advent in natural language processing (NLP) research that aimed to automatically detect users' attitudes towards a topic \citep{augenstein2016stance,derczynski2017semeval,gorrell2019semeval}. Particularly relevant to our paper, previous work studied the automatic detection of users' attitudes towards vaccination and vaccines\citep{skeppstedt-etal-2017-automatic,bello2017detecting,muller2019crowdbreaks}.
Additionally, social science research has employed survey-based methods (e.g., self-reporting questionnaires) to analyze people's stance towards vaccination on a limited scale \citep{funk2020intent,bonnevie2021quantifying}. In this section, however, we focus our review on recent papers and benchmarks related to the \textbf{COVID-19 vaccination}.

\subsection{Existing COVID-19 Vaccine Stance Dataset}
To analyze Twitter users' stance towards COVID-19 vaccination, \citet{cotfas2021longest} introduce the first open-source\footnote{Here, we only discuss publicly available datasets as we need access to the dataset specifications.} dataset containing English tweets published after the first month of the announcement of the Pfizer \& BioNTech COVID-19 vaccine (i.e., from Nov 2020 to Dec 2020). Based on the data collection pipeline provided by \citet{cotfas2021longest}, \citet{poddar2022winds} and \citet{delcea2022new} develop two extended datasets which cover a limited time span (i.e., 3-month and 1-month after Dec 2020, respectively). In addition to monolingual datasets in English, \citet{chen2022multilingual} release the first collection of multilingual tweets published in Luxembourg related to COVID-19 vaccines. However, we observe that the majority of tweets in \citet{chen2022multilingual} are in French, which is the official language of Luxembourg. In general, all datasets mentioned above are developed for a standard three-way classification experimental setup, i.e., applying supervised methods to map a tweet into one of the stance categories including \textit{anti-vaxx}, \textit{pro-vaxx}, and \textit{irrelevant}. We also observed a number of open-source unlabelled datasets \citep{deverna2021covaxxy} related to COVID-19 vaccination that is used for statistical analysis in the field of social sciences. We further display more specifications of these datasets in Table~\ref{tab:dataset}.

\subsection{COVID-19 Vaccine Stance Classification}
Given the user-generated content, previous studies have used standard supervised classifiers (such as SVM and BERT) to identify users' stance towards COVID-19 vaccination \citep{muller2019crowdbreaks,cotfas2021longest,chen2022multilingual}. Similar to \citep{gururangan2020don}, we also observe that domain adaptive pre-trained language models (PLMs) can significantly perform better than vanilla PLMs . For example, COVID-BERT \citep{muller2020covid} shows an improvement compared to the original bert-large checkpoint \citep{devlin2019bert} in various COVID-19 downstream tasks especially automatic detection of users' attitudes towards COVID-19 events \citep{muller2020covid,cotfas2021longest,poddar2022winds}. This suggests that the second phase of pre-training (i.e. the domain \& task adaptive strategy) makes the COVID-BERT to capture more temporal \& topical information for COVID-specific tasks.

As for the user-level analysis, \citet{poddar2022winds} shed light on the temporal concept drift of the percentage of users who were against the vaccine before and after COVID-19 pandemic by using tweets stance classifier to map the original publisher (Twitter users) into \textit{pro-vaxxer} or \textit{anti-vaxxer}. Besides, other studies have focused on the difference in user demographic features between people who support and those who are against COVID-19 vaccination \citep{wang2021multilevel,almadan2022will,aw2021covid}. There are several reasons for the COVID-19 vaccine hesitancy, including potential side effects, distrust of the government (health system), and the vaccine developers \citep{johnson2020online,praveen2021analyzing,germani2021anti,CAVES}. First Draft has also reported on widespread, high engagement vaccine disinformation narratives, as well as on the presence of “data deficits” where media and government sources are failing to provide the right information \citep{dodsoncovid}.\footnote{\url{https://firstdraftnews.org/long-form-article/covid-19-vaccine-research-reports/}} In particular, disinformation about vaccines is known to impact negatively on citizens trust in COVID-19 vaccination, being a direct cause of vaccine hesitancy \citep{jennings2021lack,loomba2021measuring,gisondi2022deadly}. 

\subsection{Our Work}
Previous studies have focused on analyzing user stance towards COVID-19 vaccination using standard stance categories (i.e., three-way setup). However, there is a need to distinguish between tweets that express personal concerns and hesitation about vaccination from the ones that share disinformation. For this purpose, we develop a four-category dataset enabling the analysis of vaccine hesitation on Twitter. We show details about previously published datasets and ours in Table \ref{tab:related_work}. In comparison, our dataset differs from existing ones in a number of points: (i) we further divide user stance towards COVID-19 vaccination into fine-grained categories including vaccine hesitancy which is overlooked in previous works; (ii) our dataset covers a longer temporal span (i.e., 18-month) encompassing the period from the announcement of the first COVID-19 vaccine officially administered to the second or even booster doses being made publicly available; and (iii) we propose efficient data annotation strategies that use less time and human resources.

\begin{figure}[!t]
    \centering
    \scriptsize
    \includegraphics[scale=0.25]{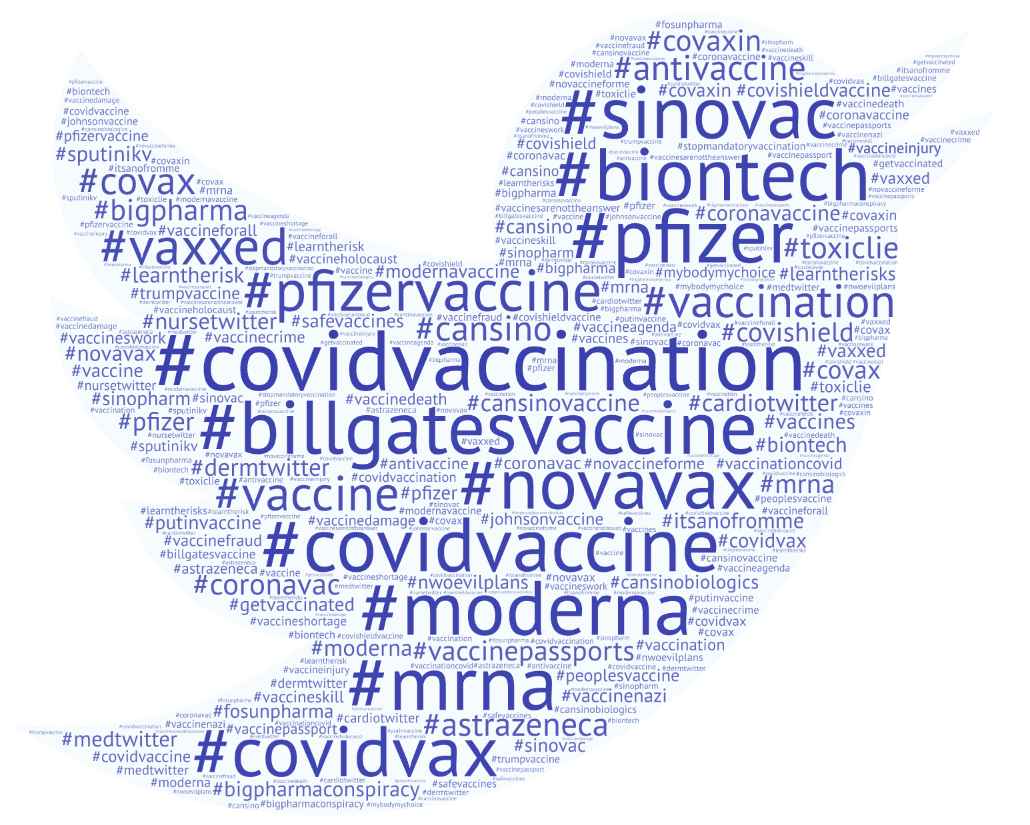}
    \caption{COVID-19 Vaccine-related Keywords.}
    \label{fig:hashtags}
\end{figure}

\begin{figure}[!t]
    \centering
    \scriptsize
    \includegraphics[scale=0.25]{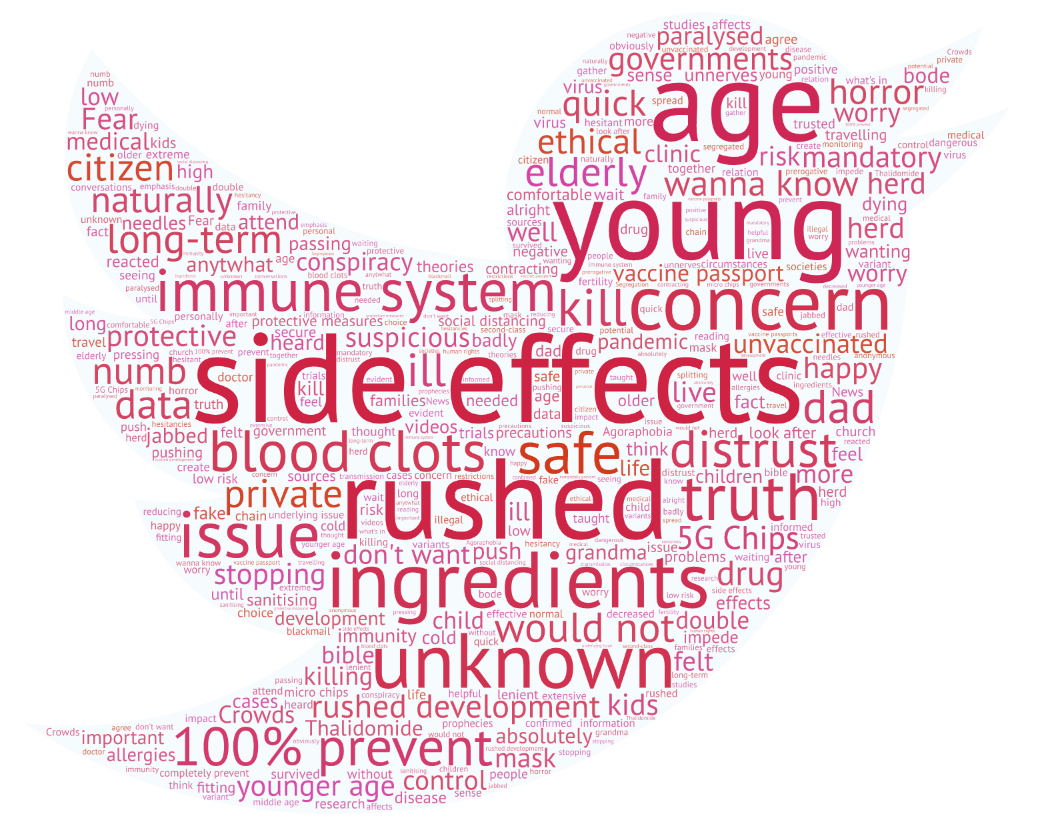}
    \caption{COVID-19 Vaccine Hesitancy Reasons.}
    \label{fig:keyworks}
\end{figure}

\section{Data}
\label{data}
In general, we frame our dataset development pipeline into three steps: 
\begin{itemize}
    \item (i) \textbf{COVID-19 Vaccine-Related Tweets Collection.} We first collect a set of COVID-19 vaccine-related tweets ($D$) through keywords searching;
    \item (ii) \textbf{Data Sampling.} Given $D$, we then filter out a representative subset of tweets $T$ to annotate;
    \item (iii) \textbf{Data Annotation.} We finally introduce the details of data annotation.
\end{itemize}

\subsection{COVID-19 Vaccine-Related Tweets Collection}
Using the streaming COVID-19 Twitter API\footnote{\url{https://developer.twitter.com/en/docs/twitter-api}}, COVID-19 vaccine-related tweets are collected automatically based on COVID-related hashtags. Following previous work \citep{cotfas2021longest,di2022vaccineu,chen2022multilingual}, our hashtags\footnote{The manually curated list of all the hashtags used for collecting the COVID-19-related tweets can be found via \url{https://github.com/GateNLP/VaxxHesitancy/blob/main/covid19_hashtags.csv}.} are  manually curated and cover a wide range of COVID-19 vaccine-related topics (see Figure~\ref{fig:hashtags}). The collected data spans from October 2020 to May 2022. The period was chosen to span the beginning of the vaccination campaign to the time when vaccination rates in the UK reached around 80\% for the first dose.\footnote{\url{https://coronavirus.data.gov.uk/details/vaccinations}} 

This yielded over 175 million tweets, which we denote as dataset $D$. It should be noted that retweets (i.e., tweets starting with `RT@') are excluded, since our aim is to collect self-expressed attitudes of Twitter users.

\subsection{Data Sampling}
The millions of tweets contained in $D$ clearly exceeded the number that could feasibly be annotated manually. Therefore, we selected a temporally and topically varied sample of tweets $T$ for annotation. 

In particular, the tweet sample $T$ was created by:
\begin{itemize}
    \item (i) \textbf{Maximizing the Temporal Span.} First, tweets in $D$ are stratified into month-long subsets, one for each month between October 2020 to May 2022 and we ensure that the final subset $T$ contains tweets from each month-long subset. This ensures that $T$ reflects the \textbf{original temporal distribution} of the 175 million vaccine-related tweets in $D$.    
  
    \item (ii) \textbf{Inclusion of Vaccine Hesitant Tweets.} 
    Since vaccine hesitant tweets are a minority class,\footnote{Prior work by \citet{chen2022multilingual} has found that only around 20\% of vaccine-related tweets in their dataset were expressing negative attitudes towards the COVID-19 vaccine and this category encompassed both anti-vaxx and vaccine-hesitant tweets.} special steps were taken to ensure that a sufficient number of such tweets was included in the sample $T$. To this end, we manually compiled a list of keywords (see Figure~\ref{fig:keyworks}) that are indicative of COVID-19 vaccine hesitancy. These were derived from a government report \footnote{\url{https://www.ons.gov.uk/}} which used a survey to collect more than 50 reasons for vaccine hesitancy, grouped into categories. The most common ones are `concern about side effects and/or long-term effects', `rushed vaccine development', `distrust of the government', etc. 
    Based on these keywords, matching tweets are extracted from each month-long set of tweets, to obtain a smaller, topically and temporally balanced subset $T$. 
    
    \item (iii) \textbf{Removing Duplicates and Highly Similar Tweets.} The previous step leaves a subset $T$ which is still too large for human annotation. Therefore, it was filtered further, to remove tweets with similar or identical textual content (i.e., topical overlaps). First, we employ topic modelling to map subset $T$ into 10k clusters, and then we extract highly frequent words for each cluster, e.g., \textit{cluster \#1 (baby, pregnant, breastfeeding, etc.)} and \textit{cluster \#2 (women, pregnant, child, etc.)}. This step allows us to remove tweets from similar topics. We also used Levenshtein Distance \cite{levenshtein1966binary} to filter out duplicates or highly similar tweets. The threshold was set to 20, which allows two or three words to be different in two tweets. 
\end{itemize}

The final subset $T$ contains 3,101 tweets in English and is of similar size to other vaccine-related  datasets (see Table~\ref{tab:related_work}). The only exception is \citet{chen2022multilingual}, which however contains only a tiny fraction of English tweets, as it is predominantly in French (over 60\%) and German (over 30\%). 

 \begin{figure}[!t]
    \centering
    \scriptsize
    \includegraphics[width=\linewidth]{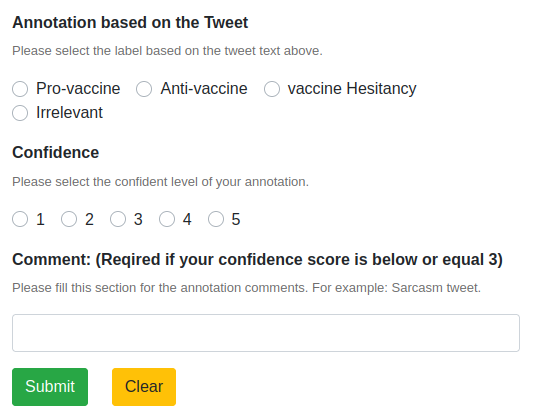}
    \caption{User Interface in GATE Teamware}
    \label{fig:pipeline}
\end{figure}

\subsection{Data Annotation}
\subsubsection{Data Annotation Workflow}
The manual data annotation workflow consisted of three separate steps, as follows: \emph{(i)} annotator training; \emph{(ii)} a quality test session, and a final \emph{(iii)} independent data annotation. All these were carried out using a collaborative web-based corpus annotation tool (GATE Teamware)\footnote{\url{https://github.com/GateNLP/gate-teamware}} \citep{bontcheva2013gate}:
\begin{itemize}
    \item (i) \textbf{Annotator Training.} First annotators were trained during in-person tutorials which introduced the GATE Teamware platform and worked through the definitions of the four vaccine stance categories, including a detailed set of real-world tweet examples (a selection of these is included in Table ~\ref{tab:examples});
    \item (ii) \textbf{Annotator Test Sessions.} To ensure that all volunteer annotators correctly aplied the annotation guidelines and produced work of good quality, a test session was organised for each annotator. This consisted of 10 tweets covering all four stance categories. A subsequent question \& answer session was also provided to explain mistakes and answer questions. In this way we ensure that annotators understand well the label definitions and can distinguish between them. 
    \item (iii) \textbf{Dataset Annotation.} Once the preparatory stages are completed, the annotators are ready to start annotating the data independently. Annotators are shown one tweet at a time and are asked to select the tweet stance, a confidence level in their annotation and an optional comment. Table \ref{tab:confidence} shows the definitions of each confidence level. Annotators can also leave an optional comment for each tweet. However, a comment is compulsory if the confidence score is three or lower. Figure~\ref{fig:pipeline} shows the corresponding GATE Teamware user interface.
\end{itemize}

\begin{table}[!t]
\small
\begin{tabular}{|l|c|l|c|c|}
\hline
\textbf{Groups} & \textbf{Annotators} & \textbf{All} & \multicolumn{1}{l|}{\textbf{Confi. \textgreater{}= 3$\dagger$}} & \multicolumn{1}{l|}{\textbf{Confi. = 5$\dagger$}} \\ \hline
\multirow{3}{*}{Group 1} & U1 \& U2   & 0.40 & 0.54 & 0.56 \\ \cline{2-5} 
                         & U1 \& U3   & 0.61 & 0.69 & 0.85 \\ \cline{2-5} 
                         & U2 \& U3   & 0.44 & 0.50 & 0.54 \\ \hline
\multirow{3}{*}{Group 2} & U4 \& U5   & 0.47 & 0.54 & 0.77 \\ \cline{2-5} 
                         & U4 \& U6   & 0.54 & 0.88 & 0.90  \\ \cline{2-5} 
                         & U5 \& U6   & 0.59 & 0.81 & 0.87 \\ \hline
\multirow{3}{*}{Group 3} & U7 \& U8   & 0.33 & 0.49 & 0.83 \\ \cline{2-5} 
                         & U7 \& U9   & 0.31 & 0.46 & 0.53 \\ \cline{2-5} 
                         & U8 \& U9   & 0.62 & 0.67 & 0.83 \\ \hline
\multirow{3}{*}{Group 4} & U10 \& U11 & 0.52 & 0.51 & 0.66 \\ \cline{2-5} 
                         & U10 \& U12 & 0.51 & 0.60 & 0.71 \\ \cline{2-5} 
                         & U11 \& U12 & 0.53 & 0.53 & 0.86 \\ \hline
\multirow{3}{*}{Group 5} & U13 \& U14 & 0.49 & 0.69 & 0.81 \\ \cline{2-5} 
                         & U13 \& U15 & 0.22 & 0.34 & 0.41 \\ \cline{2-5} 
                         & U14 \& U15 & 0.31 & 0.41 & 0.44 \\ \hline
\multirow{3}{*}{Group 6} & U16 \& U17 & 0.58 & 0.62 & 0.74 \\ \cline{2-5} 
                         & U16 \& U18 & 0.54 & 0.61 & 0.72 \\ \cline{2-5} 
                         & U17 \& U18 & 0.57 & 0.64 & 0.86 \\ \hline                    
\end{tabular}
\caption{Cohen's kappa coefficient ($K$) between every two annotators in each group. $\dagger$ denotes that the Cohen's Kappa coefficient ($K$) in column `Confi.=5' is significantly higher than the values in columns `All' and `Confi.>=3' ($t$-test, $p$ < 0.001).}
\label{IAA}
\end{table}

\subsubsection{Annotation Methodology and Quality Assurance}
Prior work has used triple annotation, where each tweet is labelled by three annotators and the final label is determined through a simple majority (i.e., a minimum of two annotators need to have selected the same label). 

Since manual annotation is time consuming and expensive, we opted for a different, more efficient strategy. 

A total of 18 volunteers were recruited to manually annotate the tweets in the subset $T$. These 18 participants were divided into six separate groups (i.e., three annotators per group). In each group 220 tweets were assigned to each annotator (660 tweets per group), which included 60 tweets also assigned to other annotators in the group. In this way we obtained 180 double annotated and 300 single-annotated tweets from each group. 

This methodology maximizes our capacity to annotate more tweets with fewer annotators. Impact on annotation quality was measured within each group by calculating the Cohen's kappa coefficient ($K$) which measures inter-rater reliability for annotator pairs (see Table~\ref{IAA}). 

Data annotation quality is improved further based on the confidence level of the tweet annotations. In particular, tweet annotations with a low confidence score were discarded. As can be seen in Table~\ref{IAA}, this improves Cohen's kappa $K$ significantly ($t$-test, $p$ < 0.05).

\begin{table*}[!ht]
\centering
\resizebox{\textwidth}{!}{%
\begin{tabular}{|c|l|}
\hline
 \textbf{Confidence} &
  \textbf{Definitions} \\ \hline 
5 &
  Extremely confident about the annotation (I’m certain about the annotation without a doubt.) \\ \hline
4 &
  \begin{tabular}[c]{@{}l@{}}Fairly confident about the annotation (I’m confident about the annotation, but might be in small \\ chance other annotators may label it in a different category.)\end{tabular} \\ \hline
3 &
  \begin{tabular}[c]{@{}l@{}}Pretty confident about the annotation (I’m pretty sure about the annotation, but might be in high \\ chance other annotators may label it in a different category.)\end{tabular} \\ \hline
2 &
  \begin{tabular}[c]{@{}l@{}}Not confident about the annotation (I’m not sure about the annotation, it seems it also belongs to \\ other categories, but you can still include this instance as a “silver standard instance” in training.)\end{tabular} \\ \hline
1 &
  \begin{tabular}[c]{@{}l@{}}Extremely unconfident about the annotation (I’m really unsure about the annotation. It may belong \\ to another category as well, you may wish to discard this instance from the training.)\end{tabular} \\ \hline
\end{tabular}%
}
\caption{Confidence Scores and Definitions}
\label{tab:confidence}
\end{table*}

\subsection{The VaxxHesitancy Dataset}

The complete VaxxHesitancy dataset has 3,101 tweets in total, including a training set with 2,670 tweets and a golden test set (double agreed) with 431 tweets (See Table~\ref{tab:dataset}).

In more detail, the VaxxHesitancy dataset is split into two parts:
\begin{itemize}  
    \item \textbf{Test set:} obtained from tweets with where both annotators agree with confidence score higher than three. 
    \item \textbf{Training set:} consists of all single annotated tweets (2,318), as they can be more noisy, and the remaining 352 double annotated tweets. For double-annotated tweets, we retain the labels from each of the annotators, their respective confidence scores, and any comments. This gives flexibility in the way this information is used during training, as discussed in our experimental Section~\ref{sec:prediction}). 
 
\end{itemize}

\begin{table}[!t]
\centering
\resizebox{\columnwidth}{!}{%
\begin{tabular}{|l|c|c|c|c|}
\hline
\rowcolor[HTML]{C0C0C0} 
\textbf{\#} & \textbf{Train 2+}  & \textbf{Train} & \textbf{Test} & \textbf{Sum} \\ \hline
Pro        & 784           & 791                & 176           & 967        \\ \hline
Anti       & 562           & 571                & 76            & 647          \\ \hline
Hesitant   & 341           & 344                & 28            & 372          \\ \hline
Irrelevant & 916           & 964                & 151           & 1,115        \\ \hline
Sum        & 2,603         & 2,670              & 431           & 3,101        \\ \hline
\end{tabular}%
}
\caption{Dataset statistics. Train set 2+ indicates that we filtered out tweets with a confidence score of 1.}
\label{tab:dataset}
\end{table}
\section{Data Characterization}
\subsection{Linguistic Analysis}
In order to investigate the differences between tweets in the \textit{anti-vaxx} and \textit{vaxx-hesitant} categories, we conduct a comparative linguistic analysis. We opted to only investigate the differences between these two categories since they were conflated in previous datasets under a common anti-vaccination category.

We perform an univariate Pearson’s correlation test to characterize which linguistic patterns
(i.e., BOW and LIWC Dictionary\footnote{We use Linguistic Inquiry and Word Count: LIWC2015 \url{https://www.liwc.app/}}) are highly correlated with each of the two categories (i.e., \textit{anti-vaxx} and \textit{vaxx-hesitant}) following the approach from \citet{Schwartz2013}. 

\subsection{BOW}
\begin{figure*}[!ht]   
    \centering
    \small
    \includegraphics[scale=0.40]{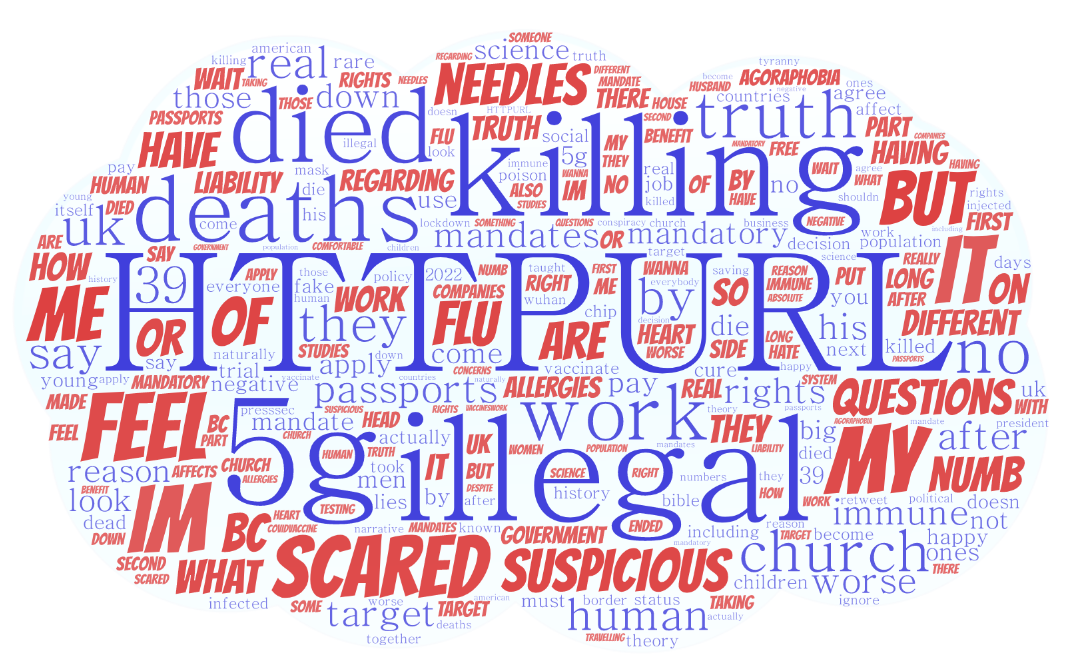}
    \caption{Top 100 BOW features associated with \textcolor{blue}{anti-vaxx} (thin) and \textcolor{red}{vacine-hesitant} (bold) categories. The larger the font, the higher the Pearson correlation value, and vice versa. HTTPURL indicates external links in anti-vaxx tweets.}
    \label{fig:bow100}
\end{figure*}

We first use the bag-of-words model (BOW) to represent each post as a TF-IDF weighted distribution over a 3,000-sized vocabulary consisting of the most frequent uni-grams (i.e., word-level tokens). We only extract tokens appearing in more than 10 and no more than 30\% of the total number of tweets. To better display the differences in BOW features associated with each category, we created a word cloud (see Figure \ref{fig:bow100}) that shows the top 100 BOW features for each of the two categories which are indicated by the different font types (i.e., \textit{\textcolor{blue}{anti-vaxx} (thin)} in blue and \textit{\textcolor{red}{vaxx-hesitant} (bold)} in red) and font size (the larger the font, the higher the Pearson correlation $r$) with the respective category.

In Figure \ref{fig:bow100} we observe that \textit{anti-vaxx} tweets contain more external links (denoted by a common HTTPURL token) and fear-inducing words such as `death', `killing', `illegal', etc. Furthermore, some of these tokens are linked to some widely spread false narratives related to the COVID-19 vaccine, e.g., `5G Micro Chip in Vaccine'\footnote{\url{https://www.politifact.com/factchecks/2021/aug/10/viral-image/link-between-covid-19-variant-and-firm-works-5g-ju/}} and `Bill Gates talked about using vaccines to control population growth'.\footnote{\url{https://www.politifact.com/factchecks/2021/oct/11/blog-posting/bill-gates-didnt-say-he-wanted-use-vaccines-reduce/}} Below we show some examples from our dataset:

\begin{quote}
    Tweet 1: \textit{`Mind Control and 5G Bill gates will insert micro chips with vaccine HTTPURL'}
\end{quote}

\begin{quote}
    Tweet 2: \textit{`The vaccine was made by the government ... and you don’t want the government to control you. Other medicines are not from the government ... and doctors are not from the government so use that'}
\end{quote}

On the other hand, tweets belonging to the \textit{vaccine-hesitant} category tend to have a more prevalent use of words related to first-person pronouns (e.g., `I'm', `me', `my', etc) and self-disclosure (e.g., `feel', `suspicious', `scared', etc.). This suggests that the vaxx-hesitant tweets in the dataset are well aligned with our definition (see Table~\ref{tab:examples}) and can be used to shed light on the posters' personal intentions for delaying or refusing the vaccine.

\begin{table}[!t]
\small
\centering
\resizebox{\columnwidth}{!}{%
\begin{tabular}{|lllc|}
\hline
\multicolumn{4}{|c|}{\textbf{LIWC}}                                                                            \\ \hline
\multicolumn{1}{|l|}{\textbf{Anti-vaxx}} & \multicolumn{1}{c|}{\textbf{$r$}} & \multicolumn{1}{l|}{\textbf{Vaccine-hesitant}} & \textbf{$r$} \\ \hline
\multicolumn{1}{|l|}{Clout}    & \multicolumn{1}{l|}{0.133} & \multicolumn{1}{l|}{1st pers singular}         & 0.294 \\ \hline
\multicolumn{1}{|l|}{Analytical thinking} & \multicolumn{1}{l|}{0.105} & \multicolumn{1}{l|}{Pronoun}   & 0.188 \\ \hline
\multicolumn{1}{|l|}{Other punctuation}   & \multicolumn{1}{l|}{0.103} & \multicolumn{1}{l|}{Authentic} & 0.187 \\ \hline
\multicolumn{1}{|l|}{All Punctuation}  & \multicolumn{1}{l|}{0.096} & \multicolumn{1}{l|}{Question marks}     & 0.185 \\ \hline
\multicolumn{1}{|l|}{Quotation marks}    & \multicolumn{1}{l|}{0.092} & \multicolumn{1}{l|}{Function}  & 0.175 \\ \hline
\multicolumn{1}{|l|}{Death}    & \multicolumn{1}{l|}{0.092} & \multicolumn{1}{l|}{Personal pronouns}     & 0.175 \\ \hline
\multicolumn{1}{|l|}{Word Count}    & \multicolumn{1}{l|}{0.089} & \multicolumn{1}{l|}{Dictionary words}       & 0.159 \\ \hline
\multicolumn{1}{|l|}{Negations}   & \multicolumn{1}{l|}{0.083} & \multicolumn{1}{l|}{Interrogatives}  & 0.132 \\ \hline
\multicolumn{1}{|l|}{Religion}    & \multicolumn{1}{l|}{0.077} & \multicolumn{1}{l|}{Prepositions}      & 0.124 \\ \hline
\multicolumn{1}{|l|}{Social}   & \multicolumn{1}{l|}{0.074} & \multicolumn{1}{l|}{Conjunctions}      & 0.124 \\ \hline
\end{tabular}
}
\caption{LIWC categories associated with anti-vaxx and vaccine hesitant tweets sorted by Pearson’s
correlation ($r$) between the normalized frequency and the labels ($p$ \textless .001).}
\label{tab:linguistic_analysis}
\end{table}

\subsection{LIWC}
We also represent each tweet using the 93 psycho-linguistic categories from the LIWC 2015 dictionary \cite{pennebaker2001linguistic}. Table~\ref{tab:linguistic_analysis} shows the top 10 most correlated LIWC categories with each of the two stance categories (anti-vaxx and hesitant).
 
The results are similar to those obtained using BOW. Namely, LIWC categories related to first-person pronouns (e.g., `1st pers singular', `Pronoun', `personal pronouns', etc) are more prevalent in tweets belonging to the \textit{vaccine-hesitant} category. Furthermore, some LIWC features such as `Interrogatives' and `Question marks' are also highly correlated with that category. This indicates that vaccine-hesitant users usually raise questions about the COVID-19 vaccine, e.g. concerning vaccine safety.

Here are two examples from the dataset:
\begin{quote}
   Tweet 3: \textit{`Would it be possible to know the ingredients of the vaccine? Just like food packaging - it is wise to know the ingredients to prevent allergies reaction.'}
\end{quote}
\begin{quote}
    Tweet 4: \textit{`\#AstraZeneca should my mom get it or not I really worry she’s 60 with asthma I really worry...'}    
\end{quote}

In comparison, we notice that frequent LIWC features in tweets belonging to the \textit{anti-vaxx} category are `Quote', `Death', `Negations', and `Religion'. This indicates that anti-vaxx tweets are more likely to use external links that refer to misleading articles or websites and/or to refer to religious reasons (e.g. `Vatican permits use of COVID-19 vaccines made using aborted fetal tissue'\footnote{\url{https://www.reuters.com/business/healthcare-pharmaceuticals/vatican-permits-use-covid-19-vaccines-made-using-aborted-foetal-tissue-2020-12-21/}}), aiming to raise fear and influence citizens against vaccination.

\begin{table*}[!t]
\large
\centering
\begin{tabular}{|lllll|}
\hline
\rowcolor[HTML]{EFEFEF} 
\multicolumn{1}{|c|}{\cellcolor[HTML]{EFEFEF}\textbf{Models}} &
  \multicolumn{1}{c|}{\cellcolor[HTML]{EFEFEF}\textbf{Accuracy}} &
  \multicolumn{1}{c|}{\cellcolor[HTML]{EFEFEF}\textbf{Precision}} &
  \multicolumn{1}{c|}{\cellcolor[HTML]{EFEFEF}\textbf{Recall}} &
  \multicolumn{1}{c|}{\cellcolor[HTML]{EFEFEF}\textbf{F1-score}} \\ \hline
\multicolumn{5}{|l|}{\textbf{All Confidence (Set 1)}} \\ \hline
\rowcolor[HTML]{EFEFEF} 
\multicolumn{1}{|l|}{\cellcolor[HTML]{EFEFEF}BERT} &
  \multicolumn{1}{l|}{\cellcolor[HTML]{EFEFEF}57.73 ± 0.77} &
  \multicolumn{1}{l|}{\cellcolor[HTML]{EFEFEF}52.08 ± 0.45} &
  \multicolumn{1}{l|}{\cellcolor[HTML]{EFEFEF}54.99 ± 0.94} &
  52.82 ± 0.47 \\ \hline
\rowcolor[HTML]{EFEFEF} 
\multicolumn{1}{|l|}{\cellcolor[HTML]{EFEFEF}COVID BERT} &
  \multicolumn{1}{l|}{\cellcolor[HTML]{EFEFEF}69.65 ± 1.43} &
  \multicolumn{1}{l|}{\cellcolor[HTML]{EFEFEF}62.70 ± 1.23} &
  \multicolumn{1}{l|}{\cellcolor[HTML]{EFEFEF}62.97 ± 1.61} &
  62.65 ± 1.40 \\ \hline
\rowcolor[HTML]{C0C0C0} 
\multicolumn{1}{|l|}{\cellcolor[HTML]{C0C0C0}VaxxBERT} &
  \multicolumn{1}{l|}{\cellcolor[HTML]{C0C0C0}72.5 ± 1.20} &
  \multicolumn{1}{l|}{\cellcolor[HTML]{C0C0C0}68.02 ± 1.24} &
  \multicolumn{1}{l|}{\cellcolor[HTML]{C0C0C0}69.35 ± 1.30} &
  68.55 ± 1.44 \\ \hline
\multicolumn{5}{|l|}{\textbf{Confidence \textgreater 1 (Set 2)}} \\ \hline
\rowcolor[HTML]{EFEFEF} 
\multicolumn{1}{|l|}{\cellcolor[HTML]{EFEFEF}BERT} &
  \multicolumn{1}{l|}{\cellcolor[HTML]{EFEFEF}58.52 ± 0.58} &
  \multicolumn{1}{l|}{\cellcolor[HTML]{EFEFEF}52.77 ± 0.63} &
  \multicolumn{1}{l|}{\cellcolor[HTML]{EFEFEF}55.53 ± 1.15} &
  53.10 ± 0.84 \\ \hline
\rowcolor[HTML]{EFEFEF} 
\multicolumn{1}{|l|}{\cellcolor[HTML]{EFEFEF}COVID BERT} &
  \multicolumn{1}{l|}{\cellcolor[HTML]{EFEFEF}69.74 ± 1.43} &
  \multicolumn{1}{l|}{\cellcolor[HTML]{EFEFEF}63.25 ± 1.09} &
  \multicolumn{1}{l|}{\cellcolor[HTML]{EFEFEF}63.65 ± 1.10} &
  63.29 ± 1.09 \\ \hline
\rowcolor[HTML]{C0C0C0} 
\multicolumn{1}{|l|}{\cellcolor[HTML]{C0C0C0}VaxxBERT} &
  \multicolumn{1}{l|}{\cellcolor[HTML]{C0C0C0}72.91 ± 1.23} &
  \multicolumn{1}{l|}{\cellcolor[HTML]{C0C0C0}68.53 ± 1.11} &
  \multicolumn{1}{l|}{\cellcolor[HTML]{C0C0C0}69.45 ± 1.71} &
  68.92 ± 1.34 \\ \hline
\multicolumn{5}{|l|}{\textbf{Higher confidence score (Set 3)}} \\ \hline
\rowcolor[HTML]{EFEFEF} 
\multicolumn{1}{|l|}{\cellcolor[HTML]{EFEFEF}BERT} &
  \multicolumn{1}{l|}{\cellcolor[HTML]{EFEFEF}59.07 ± 0.74} &
  \multicolumn{1}{l|}{\cellcolor[HTML]{EFEFEF}55.04 ± 1.04} &
  \multicolumn{1}{l|}{\cellcolor[HTML]{EFEFEF}56.51 ± 1.64} &
  54.95 ± 1.32 \\ \hline
\rowcolor[HTML]{EFEFEF} 
\multicolumn{1}{|l|}{\cellcolor[HTML]{EFEFEF}COVID BERT} &
  \multicolumn{1}{l|}{\cellcolor[HTML]{EFEFEF}71.14 ± 0.94} &
  \multicolumn{1}{l|}{\cellcolor[HTML]{EFEFEF}64.54 ± 1.29} &
  \multicolumn{1}{l|}{\cellcolor[HTML]{EFEFEF}66.61 ± 1.03} &
  65.05 ± 1.17 \\ \hline
\rowcolor[HTML]{C0C0C0} 
\multicolumn{1}{|l|}{\cellcolor[HTML]{C0C0C0}VaxxBERT} &
  \multicolumn{1}{l|}{\cellcolor[HTML]{C0C0C0}\textbf{73.04 ± 0.94}} &
  \multicolumn{1}{l|}{\cellcolor[HTML]{C0C0C0}\textbf{68.71 ± 1.59}} &
  \multicolumn{1}{l|}{\cellcolor[HTML]{C0C0C0}\textbf{70.22 ± 1.36}} &
  \textbf{69.29 ± 1.31} \\ \hline
\end{tabular}%
\caption{Model predictive performance.}
\label{tab:results}
\end{table*}

\section{COVID-19 Vaccine Stance Prediction}\label{sec:prediction}
\subsection{Baseline Models}
Following previous work \citep{cotfas2021longest,poddar2022winds,chen2022measuring}, we train two strong baselines to classify posts into our four categories:
\begin{itemize}
    \item \textbf{BERT}: Following \citet{devlin2019bert}, we directly fine-tune BERT by adding a fully connected layer on the top of the bert-large\footnote{\url{https://huggingface.co/bert-large-uncased}} model. We consider the special token (i.e., [CLS]) as the tweet-level representation. 
    \item \textbf{COVID-BERT} \citep{muller2020covid} is a domain-adapted uncased BERT model which is pretrained on COVID-19-related tweets (i.e., `covid-twitter-bert-v2').\footnote{\url{https://huggingface.co/digitalepidemiologylab/covid-twitter-bert-v2}} We fine-tune the COVID-BERT model using the same strategy as in \citet{devlin2019bert}.
\end{itemize}

\subsection{Domain Specific PLM (Our VaxxBERT Model)}
Following \citet{gururangan2020don}, we develop the  VaxxBERT domain-specific language model, which is based on the uncased COVID-BERT\citep{muller2020covid}, which has been trained on 160 million unannotated (raw) tweets related to the COVID-19 virus. 

In VaxxBERT, we continue pretraining the COVID-BERT checkpoint on over 175 million \textbf{unlabelled domain-specific tweets} (i.e., we use all remaining tweets in $D$). Following \citet{muller2020covid} and \citet{gururangan2020don}, we set the sequence length to 256, batch size to 96, validation ratio to 10\% of the training set and learning rate to 2e-5. We randomly mask tokens across epochs (2 epochs in total) with the default token masking rate (15\%). Model checkpoints are saved every 50,000 training steps and the best checkpoint is selected by the lowest validation loss. For domain-specific pretraining, we use open-source PyTorch scripts from the transformers library\footnote{\url{https://github.com/huggingface/transformers/tree/main/examples/pytorch/language-modeling}} \citep{wolf2020transformers}. VaxxBERT was trained for 240 hours on two NVIDIA GeForce RTX Graphic cards with 24GB memory.

\subsection{Experimental Setup}
All tweets are pre-processed by replacing URLs and user \texttt{@mentions} with special tokens (i.e., HTTPURL and @USER, respectively). The maximum sequence length is set to 256 tokens. We use CrossEntropyLoss as the loss function and the Adam \cite{kingma2014adam} optimizer. All models are trained using an early-stopping strategy with a learning rate $l=3e-6$, and batch size of 32. We train each model five times with different random seeds, and report the mean accuracy, precision, recall and F1-score in Table \ref{tab:results}.

\subsection{Results}
We design a battery of controlled experiments to test the quality of our dataset using the following subsets:

\begin{itemize}
    \item \textbf{Set 1} There are 352 double-annotated tweets and 2318 single-annotated tweets in the training set. We keep all single-annotated ones and randomly choose a label from two annotations for these double-annotated ones.

    \item \textbf{Set 2} We retain only tweets with annotations with confidence score of 2 or higher (285 tweets) from \textbf{Set 1}. This is motivated by the definition of confidence score of 1 , which indicates that the annotator is extremely unsure about the stance category and wishes to drop the tweet from training. In addition, all remaining double-annotated tweets have one of their two labels chosen at random as the category used for training.

    \item \textbf{Set 3} is similar to Set 2, except that the labels of all double-annotated tweets are selected based on which one of the two has a higher confidence score. In cases where both annotations have the same confidence score but different categories, then one of them is selected at random.   

 \end{itemize}

 \begin{table*}[!t]
\centering
\begin{tabular}{|c|l|c|l|}
\hline
\textbf{Column} & \textbf{Field}        & \textbf{Type} & \textbf{Description}  \\ \hline
0& id                  & str           & Unique Tweet ID                \\ \hline
1&user1\_stance1       & str           & Stance by Annotator 1          \\ \hline
2&user1\_comment       & str           & Comment by Annotator 1         \\ \hline
3&user1\_confidence    & int           & Confidence Score by Annotator 1      \\ \hline
4&user1\_time          & int           & Time (i.e., number of seconds) used by Annotator 1 \\ \hline
5&user1\_annotator\_id & str           & Annotator 1 ID                   \\ \hline
6&user2\_stance1       & str           & Stance by Annotator 2          \\ \hline
7&user2\_comment       & str           & Comment by Annotator 2         \\ \hline
8&user2\_confidence    & int           & Confidence Score by Annotator 2      \\ \hline
9&user2\_time          & int           & Time (i.e., number of seconds) used by Annotator 2 \\ \hline
10&user2\_annotator\_id & str          & Annotator 2 ID                   \\ \hline
\end{tabular}%
\caption{Dataset Columns. Note that single annotated tweets are only linked to columns from 0 to 5, the remaining columns are filled with `N/A'.} 
\label{tab:Dataset Columns}
\end{table*}

\textbf{Set 1 v.s. Set 2} We first compare the performance of models trained on all tweets in the training set (Set 1) vs those trained on Set 2, i.e. annotations with confidence scores higher than 1.  

We notice that most of the models trained on set 2 perform slightly better than those trained on set 1 according to all evaluation metrics. This indicates that dropping tweets with low confidence annotations helps improve the overall quality of both the dataset and the models.
    
\textbf{Set 2 v.s. Set 3} 
We also compare the performance of models trained on tweets where labels of double annotated tweets are either chosen at random (Set 2) or on the basis of a higher confidence score (Set 3).

In this case, models trained on Set 3 perform better than those trained on Set 2. Combined with the results above, we argue that considering annotator confidence helps to discard low-quality annotations  and helps select better labels in cases where annotators disagree. 
    
\textbf{VaxxBERT v.s. Baselines} To evaluate the performance of the domain adaptive pretrained model, we also compare VaxxBERT against the two strong baselines introduced above (i.e., bert-large and COVID-BERT). In general, our VaxxBERT model significantly (($t$-test, $p$ < 0.001)) outperforms the two baselines on all three sets (i.e., Set 1, Set 2, Set 3). Among them, VaxxBERT trained on Set 3 achieves the best performance (i.e., 73.0\% accuracy and 69.3\% F1-measure). This demonstrates that the extra domain-specific information helps VaxxBERT to improve its predictive performance, which aligns with the findings of \citet{gururangan2020don} and \citet{muller2020covid}.

\section{Conclusion}
This research was motivated by the need of media and governments to better understand the reasons behind COVID-19 vaccine hesitancy, since automatic tools for monitoring and analysis of vaccine sentiment can help policymakers design better-informed and targeted information aimed at addressing key concerns of vaccine hesitant citizens.
 
To this end, we introduced a new open-source dataset that tracks users' attitudes toward COVID-19 vaccines on Twitter. It is focused in particular on capturing vaccine hesitant tweets. Our linguistic analysis revealed significant differences between tweets belonging to the anti-vaxx and vaccine hesitant categories. Our second contribution is in using our un-annotated collection of 175 million tweets related to COVID-19 vaccination, in order to train a domain-specific PLM (VaxxBERT) that outperforms other competitive baselines on our finer-grained vaccine stance categorisation task.

In the future, we plan to enrich our dataset with multimodal information such as images and videos. Additionally, we intend to increase the number of annotated samples in our dataset by implementing advanced NLP techniques such as active learning.

\subsection{Applications}
Our work has several practical implications:
\begin{itemize}
    \item First, our dataset and the new VaxxBERT domain-specific language model can be easily reused by researchers as they are released via widely used open science platforms (i.e., Zenodo and HuggingFace \citep{wolf2020transformers}). Furthermore, our VaxxBERT model achieves the best predictive performance, which can be used as a strong baseline in future research.
    
    \item We re-frame the previous task of predicting user stance towards COVID-19 vaccination by considering a separate vaccine hesitancy category, which we demonstrate is different from anti-vaxx tweets. This can help shed light on understanding vaccine hesitancy in other cases (e.g. MMR vaccines). Also, we propose a data collection methodology that can be adopted and applied to multilingual and multi-platform datasets.
    
    \item Our dataset captures the various reasons behind vaccine hesitancy (see Figure~\ref{fig:keyworks}) and it can be used to develop interpretable classification models that generate faithful rationales.

    \item Our dataset can be repurposed for a standard three-way setting by integrating the categories of \textit{anti-vaxx} and \textit{vaxx-hesitant}. This integration will allow for cross-dataset evaluation with other benchmarks.
    
    \item Finally, our dataset can be used for qualitative research by social scientists and psychologists in order to understand better the demographic features and personality traits of vaccine hesitant and anti-vaxx users of Twitter.

\end{itemize}

\section{Dataset Availability and Ethics Statement}
Our dataset is publicly available in compliance with the FAIR principles \citep{wilkinson2016fair}:
\begin{itemize}
    \item \textbf{Findable}: Our dataset has been published in the Zenodo dataset sharing service with a unique digital object identifier (DOI: 10.5281/zenodo.7601328). We also share the VaxxBERT model via the Huggingface platform\footnote{\url{https://huggingface.co/GateNLP/covid-vaccine-twitter-bert}}.
    \item \textbf{Accessible}: Original tweets are retrievable based on their tweet IDs using the standard Twitter API\footnote{\url{https://developer.twitter.com/en/docs/twitter-api/tweets/lookup/api-reference/get-tweets-id}}. 
    \item \textbf{Interoperable}: Table~\ref{tab:Dataset Columns} summarises the dataset structure in CSV format and the description of each column (11 columns in total). CSV datasets are easily imported and processed by most widely used data processing tools. 
    \item \textbf{Re-usable}: Anyone with a Twitter developer account can re-use our dataset. Using the transformer library \citep{wolf2020transformers} researchers can also fine-tune VaxxBERT for other NLP tasks.
\end{itemize}

This work has ethical approval \#037567 from our University Research Ethics Committee. Our data collection protocol complies with the Twitter data policies for research.\footnote{\url{https://developer.twitter.com/en/products/twitter-api}} We only share tweets IDs following the Twitter API policy and replace annotator names with identifiers.

\section*{Acknowledgements}
This research is supported by a UKRI grant EP/W011212/1 and an EU Horizon 2020 grant (agreement no.871042) (“SoBigData++: European Integrated Infrastructure for Social Mining and BigData Analytics” (http://www.sobigdata.eu)). We would like to thank George Chrysostomou and all the anonymous reviewers for their valuable feedback.

\bibliography{aaai22}

\begin{thebibliography}{39}
\providecommand{\natexlab}[1]{#1}

\bibitem[{Almadan et~al.(2022)Almadan, Maher, Pereira, and
  Guo}]{almadan2022will}
Almadan, A.; Maher, M.~L.; Pereira, F.~B.; and Guo, Y. 2022.
\newblock Will You Be Vaccinated? A Methodology for Annotating and Analyzing
  Twitter Data to Measure the Stance Towards COVID-19 Vaccination.
\newblock In \emph{Future of Information and Communication Conference},
  311--329. Springer.

\bibitem[{Augenstein et~al.(2016)Augenstein, Rockt{\"a}schel, Vlachos, and
  Bontcheva}]{augenstein2016stance}
Augenstein, I.; Rockt{\"a}schel, T.; Vlachos, A.; and Bontcheva, K. 2016.
\newblock Stance Detection with Bidirectional Conditional Encoding.
\newblock In \emph{Proceedings of the 2016 Conference on Empirical Methods in
  Natural Language Processing}, 876--885.

\bibitem[{Aw et~al.(2021)Aw, Seng, Seah, and Low}]{aw2021covid}
Aw, J.; Seng, J. J.~B.; Seah, S. S.~Y.; and Low, L.~L. 2021.
\newblock COVID-19 vaccine hesitancy—A scoping review of literature in
  high-income countries.
\newblock \emph{Vaccines}, 9(8): 900.

\bibitem[{Bello-Orgaz, Hernandez-Castro, and
  Camacho(2017)}]{bello2017detecting}
Bello-Orgaz, G.; Hernandez-Castro, J.; and Camacho, D. 2017.
\newblock Detecting discussion communities on vaccination in twitter.
\newblock \emph{Future Generation Computer Systems}, 66: 125--136.

\bibitem[{Bonnevie et~al.(2021)Bonnevie, Gallegos-Jeffrey, Goldbarg, Byrd, and
  Smyser}]{bonnevie2021quantifying}
Bonnevie, E.; Gallegos-Jeffrey, A.; Goldbarg, J.; Byrd, B.; and Smyser, J.
  2021.
\newblock Quantifying the rise of vaccine opposition on Twitter during the
  COVID-19 pandemic.
\newblock \emph{Journal of communication in healthcare}, 14(1): 12--19.

\bibitem[{Chen, Chen, and Pang(2022)}]{chen2022multilingual}
Chen, N.; Chen, X.; and Pang, J. 2022.
\newblock A multilingual dataset of COVID-19 vaccination attitudes on Twitter.
\newblock \emph{Data in Brief}, 44: 108503.

\bibitem[{Chen et~al.(2022)Chen, Chen, Pang, Borga, D’Ambrosio, and
  V{\"o}gele}]{chen2022measuring}
Chen, N.; Chen, X.; Pang, J.; Borga, L.~G.; D’Ambrosio, C.; and V{\"o}gele,
  C. 2022.
\newblock Measuring COVID-19 Vaccine Hesitancy: Consistency of Social Media
  with Surveys.
\newblock In \emph{International Conference on Social Informatics}, 196--210.
  Springer.

\bibitem[{Cotfas et~al.(2021)Cotfas, Delcea, Roxin, Ioan{\u{a}}{\c{s}}, Gherai,
  and Tajariol}]{cotfas2021longest}
Cotfas, L.-A.; Delcea, C.; Roxin, I.; Ioan{\u{a}}{\c{s}}, C.; Gherai, D.~S.;
  and Tajariol, F. 2021.
\newblock The longest month: analyzing COVID-19 vaccination opinions dynamics
  from tweets in the month following the first vaccine announcement.
\newblock \emph{Ieee Access}, 9: 33203--33223.

\bibitem[{Delcea et~al.(2022)Delcea, Cotfas, Cr{\u{a}}ciun, and
  Mol{\u{a}}nescu}]{delcea2022new}
Delcea, C.; Cotfas, L.-A.; Cr{\u{a}}ciun, L.; and Mol{\u{a}}nescu, A.~G. 2022.
\newblock New Wave of COVID-19 Vaccine Opinions in the Month the 3rd Booster
  Dose Arrived.
\newblock \emph{Vaccines}, 10(6): 881.

\bibitem[{Derczynski et~al.(2017)Derczynski, Bontcheva, Liakata, Procter, Hoi,
  and Zubiaga}]{derczynski2017semeval}
Derczynski, L.; Bontcheva, K.; Liakata, M.; Procter, R.; Hoi, G. W.~S.; and
  Zubiaga, A. 2017.
\newblock SemEval-2017 Task 8: RumourEval: Determining rumour veracity and
  support for rumours.
\newblock In \emph{Proceedings of the 11th International Workshop on Semantic
  Evaluation (SemEval-2017)}, 69--76.

\bibitem[{DeVerna et~al.(2021)DeVerna, Pierri, Truong, Bollenbacher, Axelrod,
  Loynes, Torres-Lugo, Yang, Menczer, and Bryden}]{deverna2021covaxxy}
DeVerna, M.~R.; Pierri, F.; Truong, B.~T.; Bollenbacher, J.; Axelrod, D.;
  Loynes, N.; Torres-Lugo, C.; Yang, K.-C.; Menczer, F.; and Bryden, J. 2021.
\newblock CoVaxxy: A collection of English-language Twitter posts about
  COVID-19 vaccines.
\newblock In \emph{Proceedings of the International AAAI Conference on Web and
  Social Media}, volume~15, 992--999.

\bibitem[{Devlin et~al.(2019)Devlin, Chang, Lee, and
  Toutanova}]{devlin2019bert}
Devlin, J.; Chang, M.-W.; Lee, K.; and Toutanova, K. 2019.
\newblock BERT: Pre-training of Deep Bidirectional Transformers for Language
  Understanding.
\newblock In \emph{Proceedings of the 2019 Conference of the North American
  Chapter of the Association for Computational Linguistics: Human Language
  Technologies, Volume 1 (Long and Short Papers)}, 4171--4186.

\bibitem[{Di~Giovanni et~al.(2022)Di~Giovanni, Pierri, Torres-Lugo, and
  Brambilla}]{di2022vaccineu}
Di~Giovanni, M.; Pierri, F.; Torres-Lugo, C.; and Brambilla, M. 2022.
\newblock VaccinEU: COVID-19 vaccine conversations on Twitter in French, German
  and Italian.
\newblock In \emph{Proceedings of the International AAAI Conference on Web and
  Social Media}, volume~16, 1236--1244.

\bibitem[{Dodson, Mason, and Smith(2021)}]{dodsoncovid}
Dodson, K.; Mason, J.; and Smith, R. 2021.
\newblock Covid-19 vaccine misinformation and narratives surrounding Black
  communities on social media.
\newblock \emph{First Draft}.

\bibitem[{Funk and Tyson(2020)}]{funk2020intent}
Funk, C.; and Tyson, A. 2020.
\newblock Intent to get a COVID-19 vaccine rises to 60\% as confidence in
  research and development process increases.
\newblock \emph{Pew Research Center}, 3.

\bibitem[{Germani and Biller-Andorno(2021)}]{germani2021anti}
Germani, F.; and Biller-Andorno, N. 2021.
\newblock The anti-vaccination infodemic on social media: A behavioral
  analysis.
\newblock \emph{PloS one}, 16(3): e0247642.

\bibitem[{Gisondi et~al.(2022)Gisondi, Barber, Faust, Raja, Strehlow, Westafer,
  and Gottlieb}]{gisondi2022deadly}
Gisondi, M.~A.; Barber, R.; Faust, J.~S.; Raja, A.; Strehlow, M.~C.; Westafer,
  L.~M.; and Gottlieb, M. 2022.
\newblock A deadly infodemic: social media and the power of COVID-19
  misinformation.
\newblock \emph{Journal of Medical Internet Research}, 24(2): e35552.

\bibitem[{Gorrell et~al.(2019)Gorrell, Kochkina, Liakata, Aker, Zubiaga,
  Bontcheva, and Derczynski}]{gorrell2019semeval}
Gorrell, G.; Kochkina, E.; Liakata, M.; Aker, A.; Zubiaga, A.; Bontcheva, K.;
  and Derczynski, L. 2019.
\newblock SemEval-2019 task 7: RumourEval, determining rumour veracity and
  support for rumours.
\newblock In \emph{Proceedings of the 13th International Workshop on Semantic
  Evaluation}, 845--854.

\bibitem[{Gururangan et~al.(2020)Gururangan, Marasovi{\'c}, Swayamdipta, Lo,
  Beltagy, Downey, and Smith}]{gururangan2020don}
Gururangan, S.; Marasovi{\'c}, A.; Swayamdipta, S.; Lo, K.; Beltagy, I.;
  Downey, D.; and Smith, N.~A. 2020.
\newblock Don’t Stop Pretraining: Adapt Language Models to Domains and Tasks.
\newblock In \emph{Proceedings of the 58th Annual Meeting of the Association
  for Computational Linguistics}, 8342--8360.

\bibitem[{Jennings et~al.(2021)Jennings, Stoker, Bunting, Valgar{\dh}sson,
  Gaskell, Devine, McKay, and Mills}]{jennings2021lack}
Jennings, W.; Stoker, G.; Bunting, H.; Valgar{\dh}sson, V.~O.; Gaskell, J.;
  Devine, D.; McKay, L.; and Mills, M.~C. 2021.
\newblock Lack of trust, conspiracy beliefs, and social media use predict
  COVID-19 vaccine hesitancy.
\newblock \emph{Vaccines}, 9(6): 593.

\bibitem[{Johnson et~al.(2020)Johnson, Vel{\'a}squez, Restrepo, Leahy, Gabriel,
  El~Oud, Zheng, Manrique, Wuchty, and Lupu}]{johnson2020online}
Johnson, N.~F.; Vel{\'a}squez, N.; Restrepo, N.~J.; Leahy, R.; Gabriel, N.;
  El~Oud, S.; Zheng, M.; Manrique, P.; Wuchty, S.; and Lupu, Y. 2020.
\newblock The online competition between pro-and anti-vaccination views.
\newblock \emph{Nature}, 582(7811): 230--233.

\bibitem[{Karmakharm et~al.(2023)Karmakharm, Wilby, Roberts, and
  Bontcheva}]{bontcheva2013gate}
Karmakharm, T.; Wilby, D.; Roberts, I.; and Bontcheva, K. 2023.
\newblock {GATE Teamware (Version 0.1.4) [Computer software]}.
\newblock \url{https://github.com/GateNLP/gate-teamware}.

\bibitem[{Kingma and Ba(2015)}]{kingma2014adam}
Kingma, D.~P.; and Ba, J. 2015.
\newblock Adam: {A} Method for Stochastic Optimization.
\newblock In \emph{3rd {I}nternational {C}onference on {L}earning
  {R}epresentations, {ICLR} 2015}.

\bibitem[{Levenshtein et~al.(1966)}]{levenshtein1966binary}
Levenshtein, V.~I.; et~al. 1966.
\newblock Binary codes capable of correcting deletions, insertions, and
  reversals.
\newblock In \emph{Soviet physics doklady}, 707--710. Soviet Union.

\bibitem[{Loomba et~al.(2021)Loomba, de~Figueiredo, Piatek, de~Graaf, and
  Larson}]{loomba2021measuring}
Loomba, S.; de~Figueiredo, A.; Piatek, S.~J.; de~Graaf, K.; and Larson, H.~J.
  2021.
\newblock Measuring the impact of COVID-19 vaccine misinformation on
  vaccination intent in the UK and USA.
\newblock \emph{Nature human behaviour}, 5(3): 337--348.

\bibitem[{M{\"u}ller, Salath{\'e}, and Kummervold(2020)}]{muller2020covid}
M{\"u}ller, M.; Salath{\'e}, M.; and Kummervold, P.~E. 2020.
\newblock COVID-Twitter-BERT: A Natural Language Processing Model to Analyse
  COVID-19 Content on Twitter.
\newblock \emph{arXiv preprint arXiv:2005.07503}.

\bibitem[{M{\"u}ller and Salath{\'e}(2019)}]{muller2019crowdbreaks}
M{\"u}ller, M.~M.; and Salath{\'e}, M. 2019.
\newblock Crowdbreaks: tracking health trends using public social media data
  and crowdsourcing.
\newblock \emph{Frontiers in public health}, 7: 81.

\bibitem[{Muric et~al.(2021)Muric, Wu, Ferrara et~al.}]{muric2021covid}
Muric, G.; Wu, Y.; Ferrara, E.; et~al. 2021.
\newblock COVID-19 vaccine hesitancy on social media: building a public twitter
  data set of antivaccine content, vaccine misinformation, and conspiracies.
\newblock \emph{JMIR public health and surveillance}, 7(11): e30642.

\bibitem[{Pennebaker, Francis, and Booth(2001)}]{pennebaker2001linguistic}
Pennebaker, J.~W.; Francis, M.~E.; and Booth, R.~J. 2001.
\newblock Linguistic inquiry and word count: LIWC 2001.
\newblock \emph{Mahway: Lawrence Erlbaum Associates}, 71.

\bibitem[{Poddar et~al.(2022{\natexlab{a}})Poddar, Mondal, Misra, Ganguly, and
  Ghosh}]{poddar2022winds}
Poddar, S.; Mondal, M.; Misra, J.; Ganguly, N.; and Ghosh, S.
  2022{\natexlab{a}}.
\newblock Winds of Change: Impact of COVID-19 on Vaccine-related Opinions of
  Twitter users.
\newblock In \emph{Proceedings of the International AAAI Conference on Web and
  Social Media}, volume~16, 782--793.

\bibitem[{Poddar et~al.(2022{\natexlab{b}})Poddar, Samad, Mukherjee, Ganguly,
  and Ghosh}]{CAVES}
Poddar, S.; Samad, A.~M.; Mukherjee, R.; Ganguly, N.; and Ghosh, S.
  2022{\natexlab{b}}.
\newblock CAVES: A Dataset to Facilitate Explainable Classification and
  Summarization of Concerns towards COVID Vaccines.
\newblock In \emph{Proceedings of the 45th International ACM SIGIR Conference
  on Research and Development in Information Retrieval}, 3154–3164.

\bibitem[{Praveen, Ittamalla, and Deepak(2021)}]{praveen2021analyzing}
Praveen, S.; Ittamalla, R.; and Deepak, G. 2021.
\newblock Analyzing the attitude of Indian citizens towards COVID-19 vaccine--A
  text analytics study.
\newblock \emph{Diabetes \& Metabolic Syndrome: Clinical Research \& Reviews},
  15(2): 595--599.

\bibitem[{Rzymski et~al.(2021)Rzymski, Borkowski, Dr{\k{a}}g, Flisiak,
  Jemielity, Krajewski, Mastalerz-Migas, Matyja, Pyr{\'c}, Simon
  et~al.}]{rzymski2021strategies}
Rzymski, P.; Borkowski, L.; Dr{\k{a}}g, M.; Flisiak, R.; Jemielity, J.;
  Krajewski, J.; Mastalerz-Migas, A.; Matyja, A.; Pyr{\'c}, K.; Simon, K.;
  et~al. 2021.
\newblock The strategies to support the COVID-19 vaccination with
  evidence-based communication and tackling misinformation.
\newblock \emph{Vaccines}, 9(2): 109.

\bibitem[{Schwartz et~al.(2013)Schwartz, Eichstaedt, Kern, Dziurzynski,
  Ramones, Agrawal, Shah, Kosinski, Stillwell, and Seligman}]{Schwartz2013}
Schwartz, H.~A.; Eichstaedt, J.~C.; Kern, M.~L.; Dziurzynski, L.; Ramones,
  S.~M.; Agrawal, M.; Shah, A.; Kosinski, M.; Stillwell, D.; and Seligman,
  M.~E. 2013.
\newblock {Personality, Gender, and Age in the Language of Social Media: The
  Open-vocabulary Approach}.
\newblock \emph{PloS ONE}, 8(9).

\bibitem[{Skeppstedt, Kerren, and Stede(2017)}]{skeppstedt-etal-2017-automatic}
Skeppstedt, M.; Kerren, A.; and Stede, M. 2017.
\newblock Automatic detection of stance towards vaccination in online
  discussion forums.
\newblock In \emph{Proceedings of the International Workshop on Digital Disease
  Detection using Social Media 2017 ({DDDSM}-2017)}, 1--8. Taipei, Taiwan:
  Association for Computational Linguistics.

\bibitem[{van~der Linden et~al.(2021)van~der Linden, Dixon, Clarke, and
  Cook}]{van2021inoculating}
van~der Linden, S.; Dixon, G.; Clarke, C.; and Cook, J. 2021.
\newblock Inoculating against COVID-19 vaccine misinformation.
\newblock \emph{EClinicalMedicine}, 33.

\bibitem[{Wang and Liu(2021)}]{wang2021multilevel}
Wang, Y.; and Liu, Y. 2021.
\newblock Multilevel determinants of COVID-19 vaccination hesitancy in the
  United States: A rapid systematic review.
\newblock \emph{Preventive medicine reports}, 101673.

\bibitem[{Wilkinson et~al.(2016)Wilkinson, Dumontier, Aalbersberg, Appleton,
  Axton, Baak, Blomberg, Boiten, da~Silva~Santos, Bourne
  et~al.}]{wilkinson2016fair}
Wilkinson, M.~D.; Dumontier, M.; Aalbersberg, I.~J.; Appleton, G.; Axton, M.;
  Baak, A.; Blomberg, N.; Boiten, J.-W.; da~Silva~Santos, L.~B.; Bourne, P.~E.;
  et~al. 2016.
\newblock The FAIR Guiding Principles for scientific data management and
  stewardship.
\newblock \emph{Scientific data}, 3(1): 1--9.

\bibitem[{Wolf et~al.(2020)Wolf, Debut, Sanh, Chaumond, Delangue, Moi, Cistac,
  Rault, Louf, Funtowicz et~al.}]{wolf2020transformers}
Wolf, T.; Debut, L.; Sanh, V.; Chaumond, J.; Delangue, C.; Moi, A.; Cistac, P.;
  Rault, T.; Louf, R.; Funtowicz, M.; et~al. 2020.
\newblock Transformers: State-of-the-art natural language processing.
\newblock In \emph{Proceedings of the 2020 conference on empirical methods in
  natural language processing: system demonstrations}, 38--45.

\end{thebibliography}
\end{document}